%% file: main.tex
\documentclass[10pt,twocolumn,letterpaper]{article}

\usepackage{cvpr}              % To produce the CAMERA-READY version
\input{preamble}

\definecolor{cvprblue}{rgb}{0.21,0.49,0.74}
\usepackage[pagebackref,breaklinks,colorlinks,allcolors=cvprblue]{hyperref}

\def\paperID{*****} % *** Enter the Paper ID here
\def\confName{CVPR}
\def\confYear{2025}

\newcommand{\modelname}{DINeMo}
\title{\modelname: Learning Neural Mesh Models with no 3D Annotations}

\author{\textbf{Weijie Guo$^{1,2}$\thanks{Work done when W.G. was a full-time intern with JHU.} \quad Guofeng Zhang$^{1}$ \quad Wufei Ma$^{1}$ \quad Alan Yuille$^{1}$} \\
\small $^{1}$Johns Hopkins University \quad $^{2}$ Peking University
}

\begin{document}
\maketitle
\input{sec/0_abstract}    
\input{sec/1_intro}
\input{sec/2_relatedwork}
\input{sec/3_methods}
\input{sec/4_experiments}
\input{sec/5_conclusions}
{
    \small
    \bibliographystyle{ieeenat_fullname}
    \bibliography{main}
}

\input{sec/X_suppl}

\end{document}

%% file: preamble.tex
\usepackage{comment}
\usepackage{multirow}
\usepackage{colortbl}

\definecolor{mygray}{gray}{0.6}
\definecolor{mygray-bg}{gray}{0.9}

%% file: sec/0_abstract.tex
\begin{abstract}
Category-level 3D/6D pose estimation is a crucial step towards comprehensive 3D scene understanding, which would enable a broad range of applications in robotics and embodied AI. Recent works explored neural mesh models that approach a range of 2D and 3D tasks from an analysis-by-synthesis perspective. Despite the largely enhanced robustness to partial occlusion and domain shifts, these methods depended heavily on 3D annotations for part-contrastive learning, which confines them to a narrow set of categories and hinders efficient scaling.
In this work, we present \modelname, a novel neural mesh model that is trained with no 3D annotations by leveraging pseudo-correspondence obtained from large visual foundation models. We adopt a bidirectional pseudo-correspondence generation method, which produce pseudo correspondence utilizing both local appearance features and global context information.
Experimental results on car datasets demonstrate that our \modelname{} outperforms previous zero- and few-shot 3D pose estimation by a wide margin, narrowing the gap with fully-supervised methods by 67.3\%. Our \modelname{} also scales effectively and efficiently when incorporating more unlabeled images, which demonstrate the advantages over supervised learning methods that rely on 3D annotations. Our project page is available \href{https://analysis-by-synthesis.github.io/DINeMo/}{here}.
\end{abstract}

%% file: sec/1_intro.tex
\section{Introduction} \label{sec:intro}

Estimating the 3D location and 3D orientation of objects from a certain category is both a challenging and essential step towards comprehensive scene understanding~\cite{caesar2020nuscenes,ma2022robust,brazil2023omni3d,jesslen2024novum}. Models must learn to inference 3D formulations from 2D signals only, while also generalizing across different shapes and appearances within a category. However, generalization to out-of-distribution (OOD) scenarios, \eg, partial occlusions, novel shapes and appearances, remain a fundamental challenge~\cite{hendrycks2019benchmarking,kortylewski2020compositional,zhao2022ood}.

Inspired by cognitive studies on human vision~\cite{neisser1967cognitive,yuille2006vision}, recent works explored compositional 3D representations of objects, \eg, neural mesh models, and approached a range of 2D and 3D tasks with an analysis-by-synthesis process~\cite{wang2021nemo,ma2022robust,jesslen2024novum}. By learning a part-contrastive 3D feature representation of the objects, these methods can jointly optimize the 3D location, 3D orientation, and part visibility of objects with a feature construction loss. Despite the improved robustness to partial occlusion and domain shifts~\cite{zhao2022ood}, these methods require various 3D annotations that enable part-contrastive learning, such as object orientations~\cite{xiang2014beyond,ma2024imagenet3d} or human poses~\cite{zhang2023nbf,ionescu2013human3}. These 3D annotations are often hard to obtain --- they are time-consuming to annotate and require certain expertise from the annotators, which limits the applicability of these neural mesh models to a broader range of objects or to scaling up efficiently. To address the scarcity of 3D annotations, previous studies~\cite{ma2023generating} explored generative models to produce synthetic images with 3D annotations. However, models trained on synthetic data must also be finetuned on real data with 3D annotations given the considerable domain gap between diffusion-generated and real images.

In this work, we present \modelname{}, a novel neural mesh model that is trained without 3D annotations , enabled by leveraging large pretrained visual foundation models, such as DINOv2~\cite{oquab2023dinov2}. Rather than relying on groundtruth keypoint locations as in previous works~\cite{ma2022robust,jesslen2024novum}, we train \modelname{} with part-contrastive loss using pseudo-correspondence obtained by matching neural features from SD-DINO~\cite{zhang2023tale}. However, raw pseudo-correspondence from SD-DINO can be quite noisy, \eg, keypoints are often mismatched between left and right. We argue that keypoint correspondence matching should consider both local information, \ie, per patch feature similarities, and global context information, \ie, 3D orientation of the object.
Based on this motivation, we propose a novel bidirectional pseudo-correspondence generation, which consists of two steps: (i) matching a global pose label from raw keypoint correspondences, and (ii) refine local keypoint correspondences based on the predicted global pose label. Lastly we extend standard analysis-by-synthesis inference with Grounded-SAM\cite{ren2024grounded} masks, achieving enhanced occlusion robustness.

% We evaluate our \modelname{} on PASCAL3D+~\cite{xiang2014beyond} for standard 3D pose estimation and on occluded PASCAL3D+~\cite{wang2020robust} for partial occlusion generalization.
Extensive results on car class demonstrate that our \modelname{} outperforms previous zero- and few-shot 3D pose estimation methods by a wide margin on both in-distribution and partial occlusion testing data,  achieving a 67.3\% reduction in the performance gap relative to fully supervised methods. Our approach also outperforms all previous methods on SPair71k~\cite{min2019spair} for semantic correspondence.

As \modelname{} can be trained on object images without 3D annotations, we further study the scaling properties of our approach by involving abundant data from public image datasets, such as Stanford Cars~\cite{stanfordcars}. Experimental results demonstrate that our \modelname{} scales effectively and efficiently by involving more unlabeld images for training. This demonstrates the advantages of \modelname{} over previous fully-supervised approaches~\cite{ma2022robust,jesslen2024novum}, which struggle to scale up due to the reliance on 3D annotations.

In summary, our main contributions are as follows: (1) We present \modelname{}, a novel neural mesh model trained with pseudo-labels from our bidirectional pseudo-correspondence generation method. (2) Experimental results show that \modelname{} outperforms previous zero- and few-shot methods by a wide margin, largely narrowing the gap with fully-supervised methods. (3) Our \modelname{} scales effectively with more unlabeled images for training.

%% file: sec/2_relatedwork.tex
\section{Related Work} \label{sec:relatedwork}

\paragraph{Category-level pose estimation.}
Multiple strategies have been developed for category-level 3D pose estimation. Traditional methods formulate this task as a classification problem that predicts discrete pose labels \cite{tulsiani2015viewpoints, mousavian20173d}. Another line of work follows a two-stage, keypoint-driven pipeline \cite{zhou2018starmap}, where semantic keypoints are detected and used to estimate pose via Perspective-n-Point algorithm. More recent advancements shifted toward render-and-compare paradigms \cite{wang2021nemo,ma2022robust,jesslen2024novum,yang2023robust}, transitioning from generating pixel values to semantic feature representations. These methods frame pose estimation as an optimization problem that minimizes the discrepancy between features extracted from input image and ones rendered from a posed 3D mesh.

\paragraph{Keypoint correspondence.} Several approaches have been developed to establish correspondences between 2D image keypoints and 3D mesh vertices—a fundamental component in tasks such as shape reconstruction and 3D pose estimation. At the category level, however, most existing methods focused on deformable object classes, particularly humans~\cite{guler2018densepose, rempe2021humor} and animals~\cite{shtedritski2024SHIC}. This focus is largely attributed to the availability of large-scale datasets with dense image-to-template annotations, such as DensePose-COCO~\cite{guler2018densepose} for humans and DensePose-LVIS~\cite{neverova2021discovering} for animals. Notably, recent work has demonstrated the feasibility of learning dense correspondences without explicit keypoint supervision~\cite{shtedritski2024SHIC}, particularly in the animal domain. In this work, we extend this line of research to rigid object categories, which have received comparatively less attention.

\paragraph{Render-and-compare}approaches estimate 3D poses by minimizing the reconstruction error between feature representations projected from a 3D object and those extracted from the input image. This strategy can be interpreted as a form of approximate analysis-by-synthesis \cite{grenander1970unified}, which contrasts with purely discriminative methods and has demonstrated increased robustness to out-of-distribution scenarios. Particularly, such approaches have shown effectiveness in handling partial occlusions, proving beneficial in both object classification \cite{kortylewski2020combining} and 3D pose estimation tasks \cite{wang2021nemo, RePOSE}.

%% file: sec/3_methods.tex
\section{Methods}

\begin{figure}[t]
    \centering
    \includegraphics[width=0.75\columnwidth]{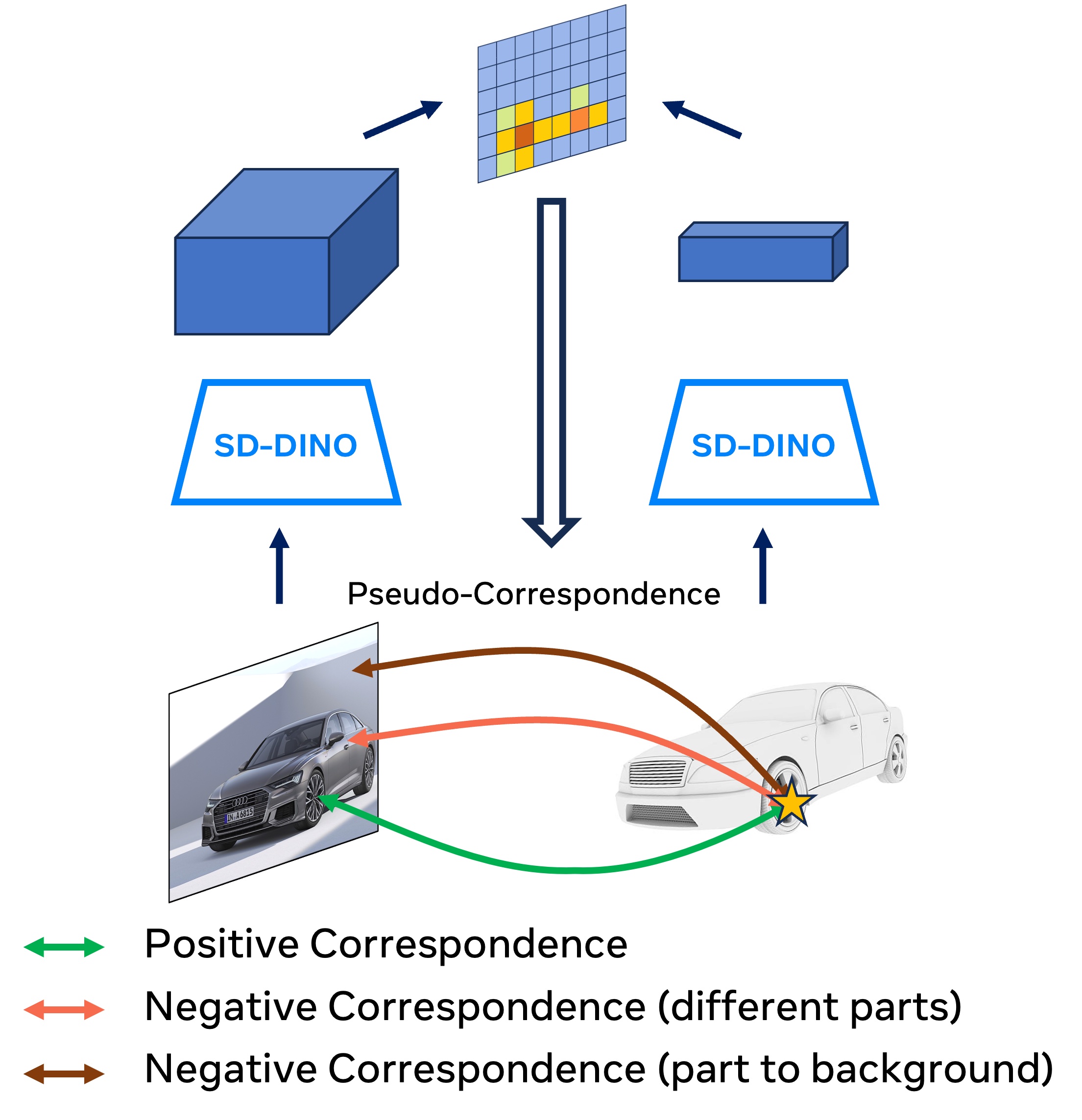}
    \vspace{-0.3cm}
    \caption{\textbf{Overview of \modelname,} a novel neural mesh model trained on pseudo-correspondence obtained from large visual foundation models.}
    \label{fig:teaser}
\end{figure}

In this section we introduce our \modelname, as illustrated in \cref{fig:teaser}. We start by reviewing the neural mesh models in \cref{sec:preliminary}. Then we introduce our bidirectional pseudo-correspondence generation in \cref{sec:pseudo}. Lastly we present our inference method in \cref{sec:inference}, which leverages pretrained segmentation model to optimize pose parameters.

\subsection{Preliminaries: Neural Mesh Models} \label{sec:preliminary}

Neural mesh models~\cite{ma2022robust,jesslen2024novum,wang2021nemo} define a probabilistic generative model $p(F \mid \mathfrak{N})$ of feature activations $F$ using a 3D neural mesh $\mathfrak{N} = \{\mathcal{V}, \mathcal{E}, \mathcal{C}\}$, where $\mathcal{V} = \{V_i \in \mathbb{R}^3\}_{i=1}^N$ is the set of mesh vertices, $\mathcal{E}$ is the edge set, and $\mathcal{C} = \{C_i \in \mathbb{R}^c\}_{i=1}^N$ is the learnable feature representation for each vertex. Given pose parameters $m$, we define the likelihood of a target feature map $F = f_\Phi(I)$ as
\begin{align}
    p(F \mid \mathfrak{N}, m, C_b) = \prod_{i \in \mathcal{FG}} p(f_i \mid \mathfrak{N}, m) \prod_{i' \in \mathcal{BG}} p(f_{i'} \mid C_b)
\end{align}
where $f_\Phi$ is the network backbone parameterized by $\Phi$, $\mathcal{FG}$ and $\mathcal{BG}$ are set of foreground and background positions, and $C_b$ is the background feature. The network parameters $\Phi$ and learnable feature $\{\mathcal{C}, C_b\}$ are optimized with a part-contrastive loss. During inference, we minimize the negative log-likelihood \wrt pose parameters $m$.

\subsection{Bidirectional Pseudo-Correspondence} \label{sec:pseudo}

\begin{figure}[t]
    \centering
    \includegraphics[width=\columnwidth]{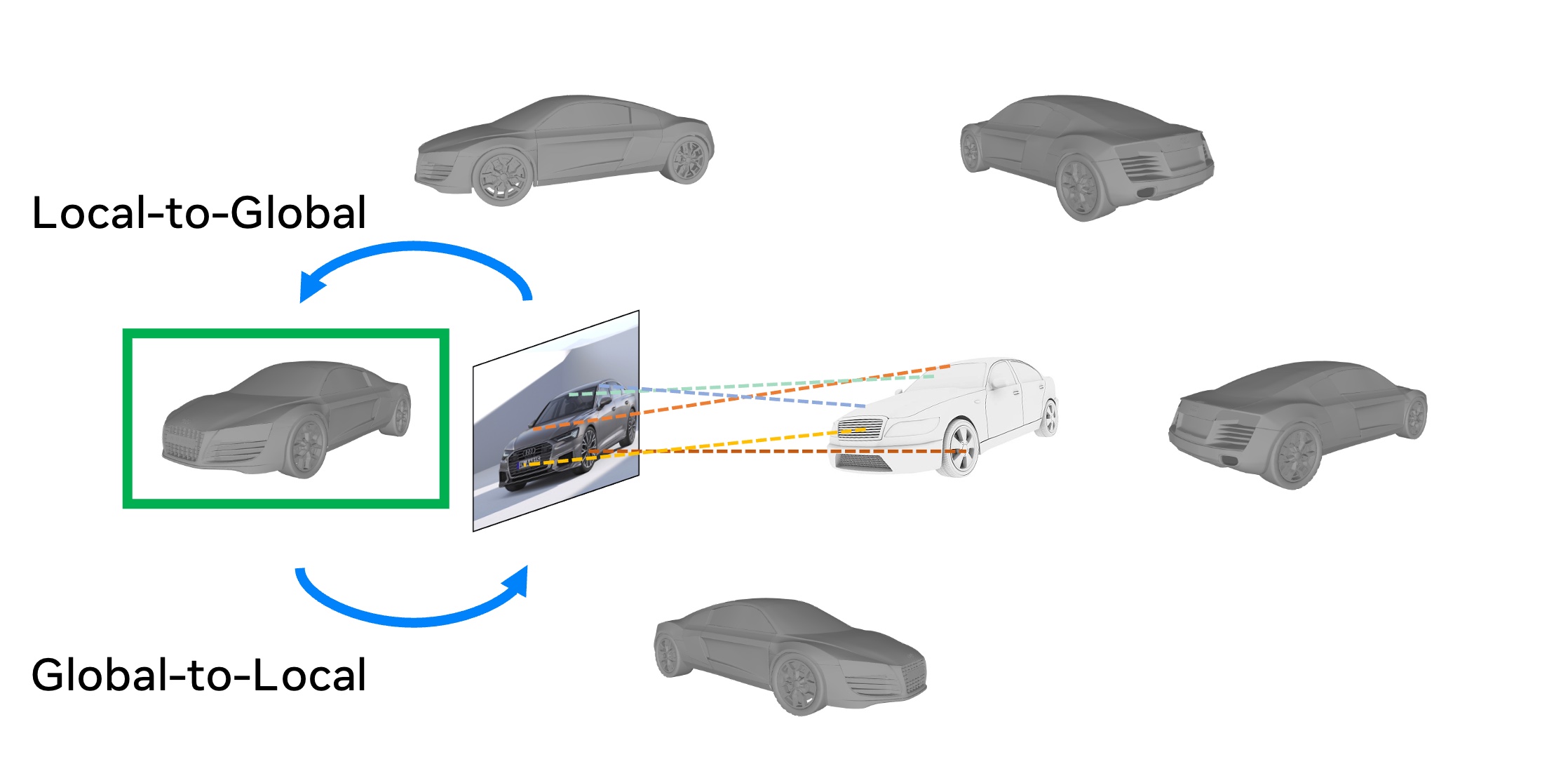}
    \vspace{-0.8cm}
    \caption{\textbf{Bidirectional pseudo-correspondence generation.} See \cref{sec:pseudo}.}
    \label{fig:bidirectional}
\end{figure}

Previous works train neural mesh models on keypoint correspondences obtained from 3D annotations, such as object or human poses. In this work, we propose to leverage pseudo-correspondence from visual foundation models, eliminating the need for 3D annotations that are difficult to obtain.

As demonstrated in \cref{fig:teaser}, we follow SD-DINO~\cite{zhang2023tale} and compute feature representations of the image and template mesh renderings from multiple views. Specifically, we extract and concatenate the DINOv2~\cite{oquab2023dinov2} features and Stable Diffusion~\cite{rombach2022high} features. We aggregate the per-vertex feature similarities from each view with the max operator and obtain the keypoint pseudo-correspondence from cosine similarities between normalized feature representations.

\begin{figure}[t]
    \centering
    \includegraphics[width=0.9\columnwidth]{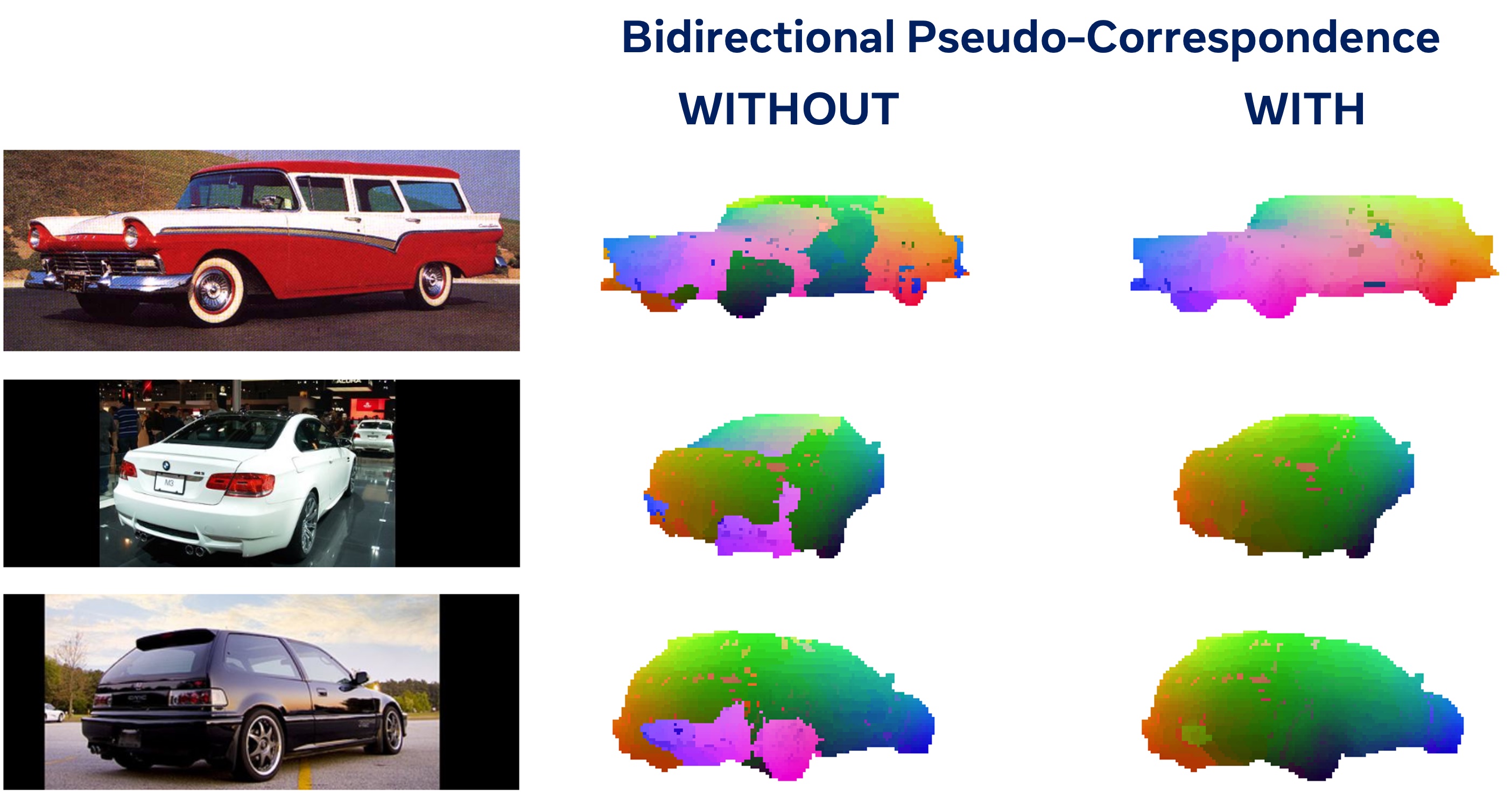}
    \vspace{-0.2cm}
    \caption{\textbf{Qualitative comparisons with and without our bidirectional pseudo-correspondence generation.} See \cref{sec:pseudo}.}
    \label{fig:inferred_pose}
\end{figure}

However, we found that raw pseudo-correspondences estimated from SD-DINO~\cite{zhang2023tale} feature similarities are noisy (see \cref{fig:inferred_pose}). In particular, object parts can be confused between left and right. This is because pseudo-correspondences from feature similarities only consider local apperances and lack of high-level consistency, \eg, 3D object pose. We argue that keypoint correspondence matching should consider both local information, such as local appearances, as well as global context information, \ie, 3D formulation of the object.

We propose \textbf{bidirectional pseudo-correspondence generation}, which generates keypoint corresopndence considering both low-level and high-level semantics. Our method consists of two steps: (i) Local-to-global: in the first stage we obtain raw pseudo-correspondences from SD-DINO~\cite{zhang2023tale} and then determine the 3D object orientation by majority voting. (ii) Global-to-local: in the second stage we refine the raw pseudo-correspondences by downweighting matching scores with vertices that are not visible from the estimated 3D orientation by a fixed constant. Our approach effectively integrates both low-level feature similarities and high-level context information and generates more consistent keypoint pseudo-correspondence as shown in \cref{fig:inferred_pose}.

% \subsection{Neural Mesh Training} \label{sec:training}

% \textcolor{red}{Not finished}

% While this method can yields satisfactory correspondences for individual images, we attempt to use this as training data

% \begin{enumerate}
%     \item Multiply DINOv2 s14 image features with feature bank to get similarity; 
%      \item Calculate CE loss between similarity and pseudo correspondence;
%      \item Backpropagate. Currently, we just use backpropagate to update the feature bank instead of momentum update.
% \end{enumerate}

\subsection{Inference} \label{sec:inference}

During inference, we minimize the negative log-likelihood \wrt pose parameters $m$ with gradient descent
\begin{align}
    & \mathcal{L}_\text{NLL}(F, \mathfrak{N}, m, C_b) \nonumber \\
    = & -\ln p(F \mid \mathfrak{N}, m, C_b) \nonumber \\
    = & -\sum_{i \in \mathcal{FG}} \ln p(f_i \mid \mathcal{N}, m) - \sum_{i' \in \mathcal{BG}} \ln p(f_{i'} \mid C_b)
\end{align}
To find foreground and background regions, previous works~\cite{ma2022robust,jesslen2024novum,wang2021nemo} introduce an one-hot map $\mathcal{Z}$ indicating if each 2D location is visible or not based on feature activations. Empirically we find that the one-hot map often involves a significant portion of the background, leading to degraded performance on partial occlusion data. Given the recent advancements of segmentation methods, we first predict $\mathcal{Z}$ with a SAM2~\cite{ravi2024sam2} model and then optimizes the negative log-likelihood loss while fixing $\mathcal{Z}$.

% At inference time, we pass the image through our backbone and compare the extracted features with our feature bank to determine the corresponding vertex for each pixel. To ensure computations focus only on the target object, we use masks obtained via the Grounded-SAM\cite{ren2024grounded} model. With our image feature extractor and neural mesh, we can extend this to other downstream tasks: 

% \textbf{Object Pose estimation}. As studied extensively in previous works\cite{wang2021nemo}, we can render neural mesh into feature maps (vertex features can be viewed as neural textures), and solve the optimal pose via render-and-compare. Specifically, we first render a number of initial poses, and select the best match with the image feature map. Then, we do gradient descent on the pose parameters to refine the coarse pose.

% \textbf{Semantic Correspondence}. We can establish correspondences in two ways: (1) directly performing a nearest-neighbor search between feature map pairs using our image feature extractor or (2) first determining the corresponding vertex of the source point and then identifying the best-matching point on the target image. Previous work has shown that the second approach yields superior performance in animal categories\cite{shtedritski2024SHIC}.

%% file: sec/4_experiments.tex
\section{Experiments}

In this section we present our experiments results on 3D pose estimation and semantic correspondence in \cref{sec:results}. We study the scaling properties of our \modelname{} in \cref{sec:scaling}. For details about our experimental setup, please refer to \cref{sec:exp_setup} in supplementary materials.

\subsection{Main Results} \label{sec:results}

\begin{table*}[ht]
\centering
\resizebox{1\textwidth}{!}{\begin{tabular}{lcccccccccccccccc}
\toprule
\multirow{2.5}{*}{{Methods}} & \multicolumn{2}{c}{L0} & & \multicolumn{2}{c}{L1} & & \multicolumn{2}{c}{L2} & & \multicolumn{2}{c}{L3} \\
\cmidrule{2-3} \cmidrule{5-6} \cmidrule{8-9} \cmidrule{11-12}
              & Acc@${\frac{\pi}{6}}$ & Acc@${\frac{\pi}{18}}$ && Acc@${\frac{\pi}{6}}$ & Acc@${\frac{\pi}{18}}$  && Acc@${\frac{\pi}{6}}$ & Acc@${\frac{\pi}{18}}$ && Acc@${\frac{\pi}{6}}$ & Acc@${\frac{\pi}{18}}$ \\ \midrule
\textbf{\textit{Fully-Supervised}} \\
Resnet50~\cite{he2016deep}   &  95.5 &   63.5    &&  80.0 &  40.7 &&  57.0  &  21.4  && 36.9 & 7.6 \\
NOVUM~\cite{jesslen2024novum}         &  97.9    &    94.9    &&   91.9   &   78.0 &&  77.1 &  52.3   &&   49.8   &  23.8  \\ \midrule
\multicolumn{2}{l}{\textbf{\textit{Zero- and Few-Shot}}} \\
NVS~\cite{wang2021neural} (7-shot) & 63.8 & 36.4 && - & - && - & - && - & - \\
NVS~\cite{wang2021neural} (50-shot) & 65.5 & 39.8 && - & - && - & - && - & - \\
3D-DST~\cite{ma2023generating} (0-shot) & 82.3 & 65.4 && - & - && - & - && - & - \\
\cellcolor{mygray-bg}{\modelname{} (ours) (0-shot)} &  \cellcolor{mygray-bg}{\textbf{92.8}}                    &      \cellcolor{mygray-bg}{\textbf{78.6}}           &\cellcolor{mygray-bg}{}&    \cellcolor{mygray-bg}{87.9}                   &      \cellcolor{mygray-bg}{68.1}                 &\cellcolor{mygray-bg}{}&      \cellcolor{mygray-bg}{73.7}                 &     \cellcolor{mygray-bg}{51.5}      &\cellcolor{mygray-bg}{}& \cellcolor{mygray-bg}{43.9} & \cellcolor{mygray-bg}{23.1}  \\ \bottomrule
\end{tabular}}
\vspace{-0.2cm}
\caption{\textbf{3D object pose estimation on the car split of Pascal3D+~\cite{xiang2014beyond} and occluded PASCAL3D+~\cite{wang2020robust}.} Our \modelname{} outperforms previous zero- and few-shot 3D pose estimation methods by a wide margin, narrowing the gap with fully-supervised methods by 67.3\%.}
\label{tab:pose}
\end{table*}

\begin{table}[t]
\small
\centering
\begin{tabular}{lc}
\toprule
Methods                      & PCK@0.1     \\
\midrule
DINOv2-ViT-S/14\cite{oquab2023dinov2}                   &  48.4         \\
DINOv2-ViT-B/14\cite{oquab2023dinov2}                   &  52.8         \\
DIFT   \cite{tang2023emergent}                     &  48.3         \\
SD-DINO\cite{zhang2023tale}                           &  53.8         \\
\textcolor{mygray}{Telling Left from Right \cite{Zhang_2024_CVPR}${}^\dag$}          &  \textcolor{mygray}{60.8}         \\
\cellcolor{mygray-bg}{\modelname{} (ours)}               &  \cellcolor{mygray-bg}{\textbf{59.1}}         \\
\bottomrule
\end{tabular}
\vspace{-0.2cm}
\caption{\textbf{Semantic correspondence evaluation on car split of SPair71k~\cite{min2019spair}.} The metric is \textit{per point} PCK, following previous works~\cite{zhang2023tale,Zhang_2024_CVPR}. Our \modelname{} outperforms all previous methods by a wide margin and achieves comparable performance with \cite{Zhang_2024_CVPR} that use index to flip source keypoints at test time.}
\end{table}

\paragraph{3D Object Pose Estimation}

We evaluate our model on the car split of the PASCAL3D+ dataset~\cite{xiang2014beyond} for in-distribution 3D pose estimation and occluded PASCAL3D+ dataset~\cite{wang2020robust} for partial occlusion generalization. As we can see from the results in \cref{tab:pose}, as a zero-shot method, our \modelname{} outperforms previous zero- and few-shot methods by a wide margin, \ie, by 27.3\% over NVS~\cite{wang2021neural} (50-shot) and by 10.5\% over 3D-DST~\cite{ma2023generating}. Moreover, our \modelname{} significantly narrows the gap between zero- and few-shot methods and fully-supervised methods by 67.3\% (from previous 15.6\% to 5.1\%). Lastly our \modelname{} demonstrate enhanced robustness to partial occlusion, largely outperforming fully-supervised ResNet50 baseline and falling behind fully-supervised NOVUM only by a small gap.

\paragraph{Semantic Correspondence}

% \textcolor{orange}{Weijie: summarize the results in table 2.}

We evaluate our model on the car split of the SPair71k dataset~\cite{min2019spair} using the PCK@0.1 metric. Results show that our model achieve a significant improvement over previous works, \eg by 10.7\% compared to the DINOv2-ViT-S/14 backbone that our method builds on and by 5.3\% compared to SD-DINO~\cite{zhang2023tale}. Our method also achieves comparable performance with Telling Left from Right~\cite{Zhang_2024_CVPR}, which utilize extra information to flip source keypoints at test time.

\subsection{Scaling Properties} \label{sec:scaling}

\begin{figure}[t]
    \centering
    \includegraphics[width=\columnwidth]{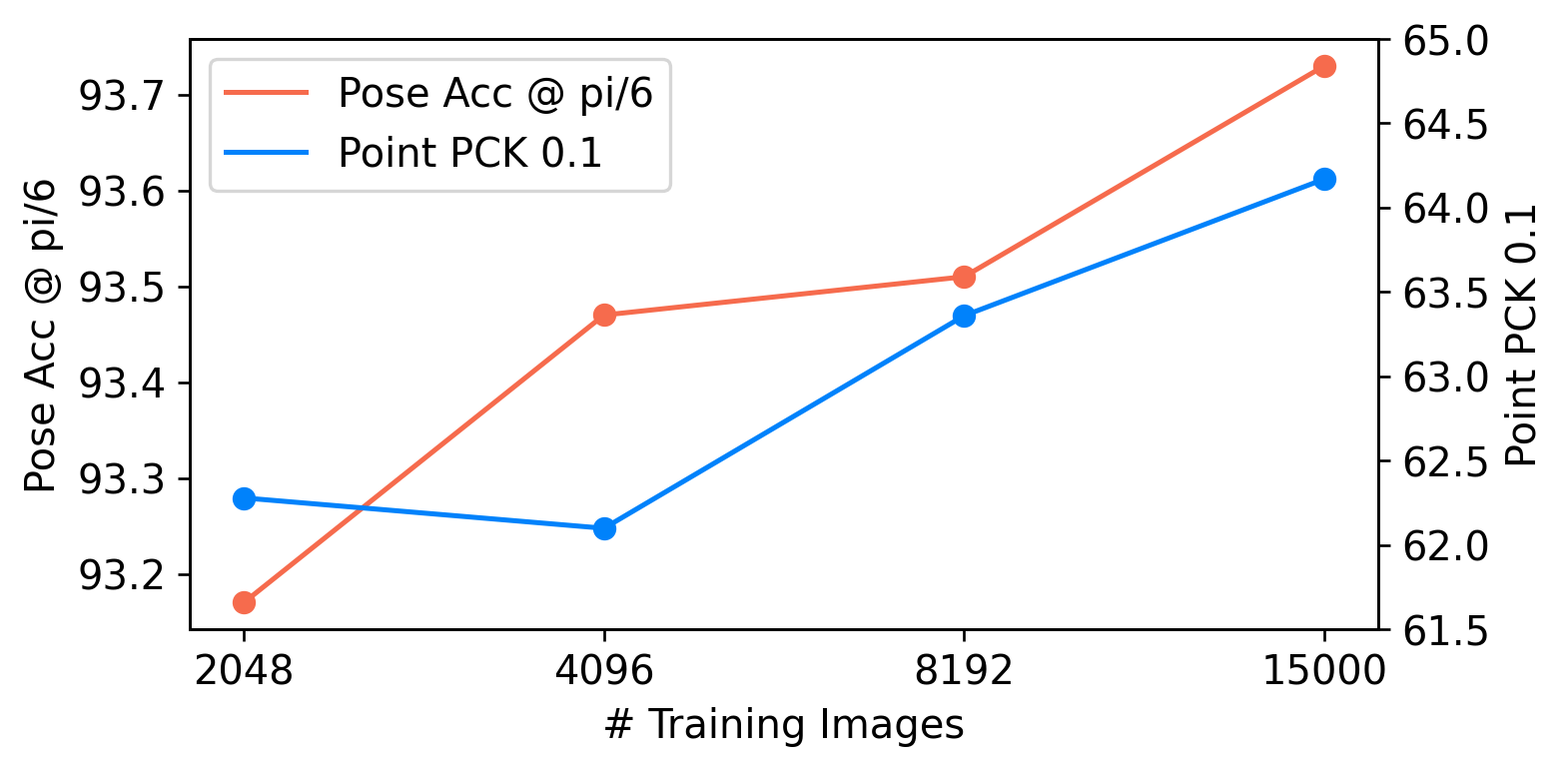}
    \vspace{-0.8cm}
    \caption{\textbf{Scaling properties of \modelname.} See \cref{sec:scaling}.}
    \label{fig:scaling}
\end{figure}

As our \modelname{} is trained on object images without 3D annotations, we study the scaling properties \wrt different training data sizes. Specifically, we trained a variety of \modelname{} models on different numbers of unlabeled images, ranging from 2048 images (comparable to the training set size in PASCAL3D+~\cite{xiang2014beyond}) up to 15,000 images (as in Stanford Cars dataset~\cite{stanfordcars}). As shown from the results in \cref{fig:scaling}, \modelname{} scales well with more unlabeled images used during training, \ie, the pose accuracy at $\pi/6$ increases from 93.1\% to 93.7\% and the per point PCK increases from 62.3 to 64.2. This highlights the advantages over previous supervised learning methods --- \modelname{} does not require 3D annotations that are hard to obtain, and scales effectively and efficiently by involving more unlabeled images for training, which are abundant on the Internet~\cite{schuhmann2022laion}.

\begin{comment}
\subsection{Scaling Properties} \label{sec:scaling}
\begin{table*}[ht]
\centering
\begin{tabular}{l|ccccc}
\hline
                    & $Acc_{\frac{\pi}{6}}$ & $Acc_{\frac{\pi}{18}}$ & MedErr & Per Image PCK 0.1 & Per Point PCK 0.1 \\ \hline
2048 Images         & 93.17           & 79.78     & 16.86      & 51.74    & 62.28             \\
4096 Images         & 93.47           & 76.53     & 16.77      & 48.51    & 59.10             \\
8192 Images         & 92.51           & 72.61     & 16.80      & 50.85    & 63.36             \\ 
15000 Images        & 93.73           & 77.34     & 15.84      & 53.80    & 64.17            \\ \hline

\end{tabular}
\caption{Performance of our model trained on Stanford Cars Dataset and tested on Pascal3D Plus}
\end{table*}
\end{comment}

% \subsection{Qualitative Comparisons} \label{sec:qualitative}

%% file: sec/5_conclusions.tex
\section{Conclusions} \label{sec:conclusions}

In this work we present \modelname{}, a novel neural mesh model that is trained with no 3D annotations-only from pseudo-correspondence obtained from large visual foundation models. We propose a novel bidirectional pseudo-correspondence generation method that can effectively utilize both local appearance features and global context information to produce more 3D-consistent pseudo-correspondence. Experimental results on car datasets demonstrate that our \modelname{} outperforms previous zero- and few-shot 3D pose estimation methods by a wide margin, narrowing the gap with fully-supervised methods by 67.3\%. By incorporating more unlabeled images during training, our \modelname{} also scales effectively and efficiently, demonstrating the advantages over fully supervised methods that rely on scarce 3D annotations.

\section{Acknowledgements}

AY acknowledges support via Army Research Laboratory award W911NF2320008 and Office of Naval Research with award N000142412696.

%% file: sec/X_suppl.tex
\clearpage
\setcounter{page}{1}
\maketitlesupplementary

\section{Author Contribution Statement}

WG conducted the experiments. WM and GZ led the study and wrote the paper. AY provided high-level conceptual guidance and feedback.

\section{Experimental Setup} \label{sec:exp_setup}

\paragraph{Benchmarks} For 3D pose estimation, we follow previous works~\cite{ma2022robust,jesslen2024novum} and evaluate our \modelname{} on the car split of PASCAL3D+~\cite{xiang2014beyond} for in-distribution testing and on the car split of occluded PASCAL3D+~\cite{wang2020robust} for partial occlusion generalization. For semantic correspondence, we evaluate our \modelname{} on the car split of the SPair71k dataset\cite{min2019spair}.

\paragraph{Baselines} For 3D pose estimation, we consider two types of baselines: (i) \textit{fully-supervised models}, \eg, ResNet50~\cite{he2016deep} and NOVUM~\cite{jesslen2024novum}, which serve as a reference to the state-of-the-art performance by utilizing all 3D pose annotations in PASCAL3D+~\cite{xiang2014beyond}; (ii) \textit{zero- and few-shot models}, \ie, NVS~\cite{wang2021neural} and 3D-DST~\cite{ma2023generating}. Specifically, NVS advanced label efficient training by synthesizing feature maps from novel views and obtain correspondences for unlabeled images. 3D-DST proposed to synthesize images with 3D annotations using diffusion models, improving zero-shot performance and enhancing model robustness.For semantic correspondence, we consider DINOv2\cite{oquab2023dinov2} and Diffusion features\cite{tang2023emergent}, as well as the unsupervised version of SD-DINO\cite{zhang2023tale} and Telling Left from Right\cite{Zhang_2024_CVPR}.

\section{Qualitative Examples} \label{sec:qualitative}

\begin{figure*}[t]
    \centering
    \includegraphics[width=\textwidth]{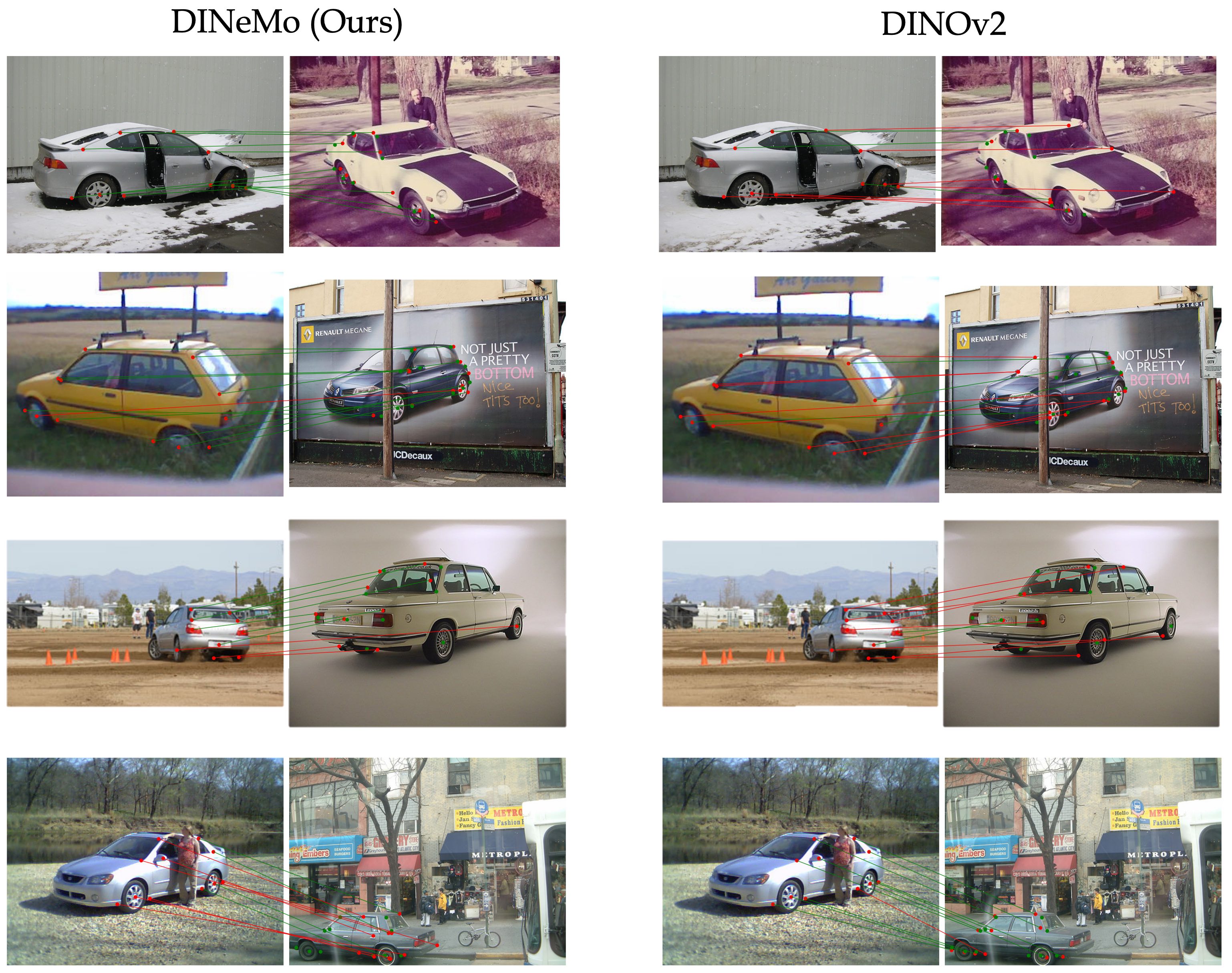}
    \vspace{-0.2cm}
    \caption{\textbf{Qualitative comparisons between DINOv2 (left) and our \modelname{} (right) on the SPair71k\cite{min2019spair} dataset.}}
    \label{fig:qualitative_correspondence}
\end{figure*}

\begin{figure*}[t]
    \centering
    \includegraphics[width=\textwidth]{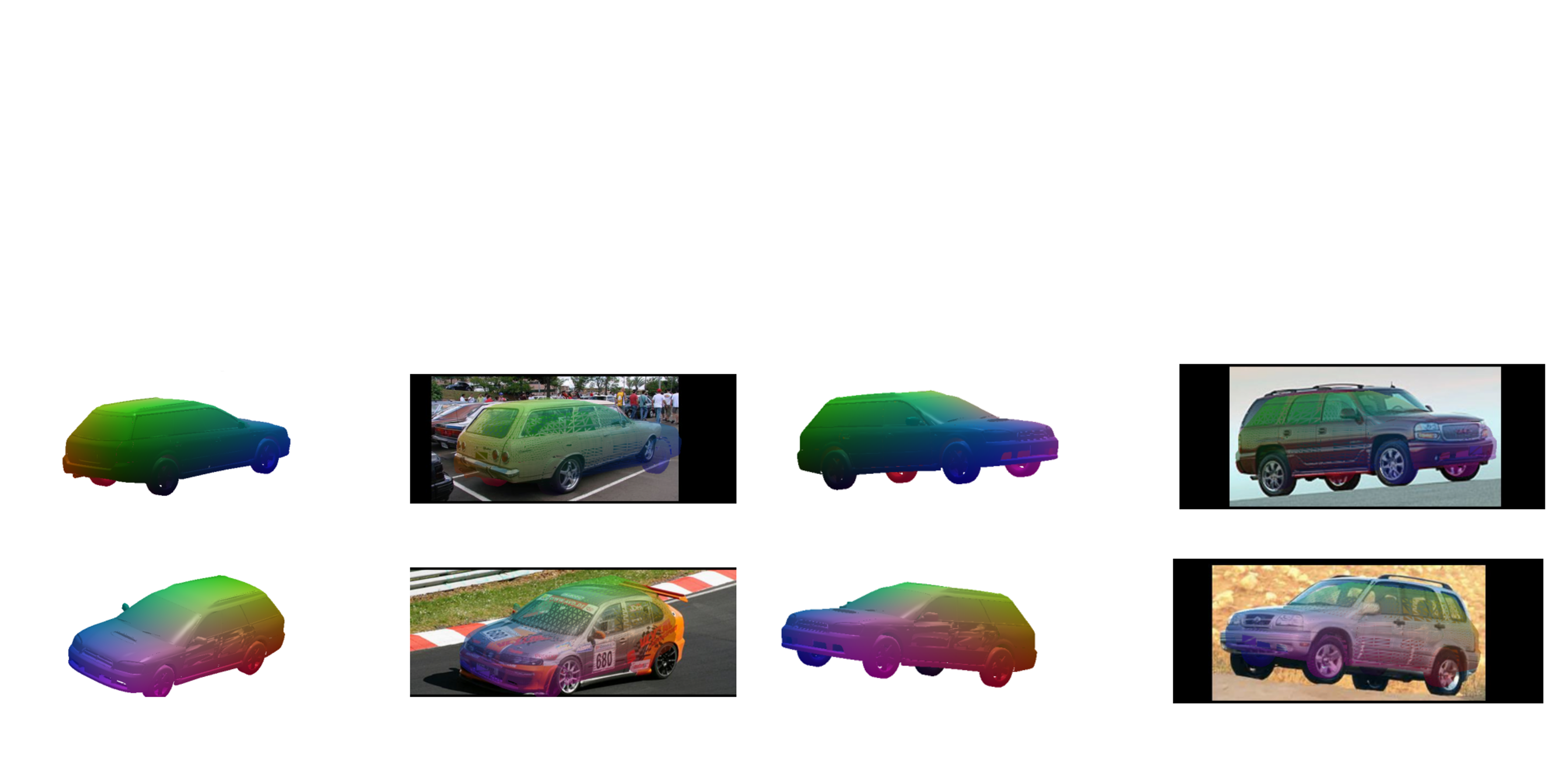}
    \vspace{-0.2cm}
    \caption{\textbf{Qualitative pose estimation results on the Pascal3D+\cite{xiang2014beyond} dataset.}}
    \label{fig:qualitative_pose}
\end{figure*}

We present some qualitative comparisons between DINOv2 and our \modelname{} in \cref{fig:qualitative_correspondence}. We also present some qualitative examples of 3D pose estimation of our \modelname{} in \cref{fig:qualitative_pose}.

\section{Public Release}

\paragraph{Code.} All code of our \modelname{} will be made available upon acceptance of the paper.

\clearpage

